\documentclass[conference,a4paper]{IEEEtran}
\IEEEoverridecommandlockouts

\usepackage[hidelinks]{hyperref}
\usepackage[cmex10]{amsmath}%American Math Society(AMS) math formatting
\usepackage{amssymb,amsfonts}%AMS extra symbols and fonts
\usepackage{dblfloatfix}%fix double column figure ordering and placement

\usepackage[ruled,vlined]{algorithm2e}
\usepackage{graphicx}
\graphicspath{{Figures/PDF/}{Figures/PNG/}}

\usepackage{booktabs}
\usepackage{siunitx}
\usepackage[numbers,compress]{natbib}
\usepackage{texnames}
\usepackage{bm,bbm}
\usepackage{orcidlink}

\usepackage{standalone} % Allows you to \input standalone files
\usepackage{tikz}
\usetikzlibrary{3d, calc, positioning, shapes.geometric, arrows.meta, backgrounds, shadows.blur}

\definecolor{inputblue}{RGB}{200, 220, 255}
\definecolor{embedpurple}{RGB}{180, 160, 220}
\definecolor{processgreen}{RGB}{160, 220, 160}
\definecolor{firecenter}{RGB}{255, 255, 100}
\definecolor{fireouter}{RGB}{220, 50, 20}
\definecolor{darkfire}{RGB}{80, 20, 20}

\makeatletter
\newsavebox{\@brx}
\newcommand{\llangle}[1][]{\savebox{\@brx}{\(\m@th{#1\langle}\)}%
  \mathopen{\copy\@brx\kern-0.5\wd\@brx\usebox{\@brx}}}
\newcommand{\rrangle}[1][]{\savebox{\@brx}{\(\m@th{#1\rangle}\)}%
  \mathclose{\copy\@brx\kern-0.5\wd\@brx\usebox{\@brx}}}
\makeatother

\begin{document}

\title{\uppercase{\LaTeX\ Template for IGARSS 2026 Articles (with \BibTeX)}
% \thanks{Identify applicable funding agency here. If none, delete this.}
}

%%%%%%%%% TITLE - PLEASE UPDATE
% \title{\uppercase{Learning Prescribed Fire Dynamics: Spatiotemporal Neural Models for Fuel Density Prediction}}
\title{\uppercase{Physics-Guided Spatiotemporal Neural Models for \\Fuel Density Prediction}}

%%%%%%%%% AUTHORS - PLEASE UPDATE
\author{Tolga Caglar, \quad Jaynil Jaiswal, \quad Saqib Azim, \quad Yudhir Gala, \quad Mai H. Nguyen, \quad Ilkay Altintas \\
San Diego Supercomputer Center \\ 
University of California, San Diego\\

% \author{	\IEEEauthorblockN{Tolga Caglar\orcidlink{0000-0002-8002-5341}}
% 	\IEEEauthorblockA{\textit{Victoria University of Wellington}\\
% 		6140 Wellington, New Zealand\\
% 		alejandro.frery@vuw.ac.nz}
% 	\and
% 	\IEEEauthorblockN{Hui Zhang\orcidlink{0000-0002-5283-7350}}
% 	\IEEEauthorblockA{\textit{Inner Mongolia University}\\
% 		010021 Hohhot, China\\
% 		hui.zhang@imu.edu.cn}
% 	\and
% 	\IEEEauthorblockN{Andrea Rey\orcidlink{0000-0002-9185-1382}}
% 	\IEEEauthorblockA{\textit{Universidad Nacional de Hurlingham}\\
% 		1688 República Argentina\\
% 		andrea.rey@unahur.edu.ar}
% }

% Institution1 address\\
{\tt\small \{tcaglar, mhnguyen\}@ucsd.edu}
% For a paper whose authors are all at the same institution,
% omit the following lines up until the closing ``}''.
% Additional authors and addresses can be added with ``\and'',
% just like the second author.
% To save space, use either the email address or home page, not both
% \and
% Second Author\\
% Institution2\\
% First line of institution2 address\\
% {\tt\small secondauthor@i2.org}
}

\maketitle

\begin{abstract}

This paper presents a physics-guided machine learning (PGML) framework for fuel density prediction, integrating physics constraints and domain knowledge into deep learning models to enhance model accuracy and stability. We explore three deep learning architectures — ConvLSTM, Adaptive Fourier Neural Operator (AFNONet), and Video Vision Transformer (ViViT) — to model the spatiotemporal evolution of fuel density. 
Our approach incorporates differentiable physics-informed terms in the loss function, including a mass-conserving fuel transport term and a rate-of-spread estimation.
%Our approach incorporates differentiable physical constraints, including a mass-conserving fuel transport term and a rate-of-spread estimation, to penalize physically inconsistent behavior. 
Experimental results, averaged across multiple independent trials, demonstrate that the proposed PGML framework outperforms purely data-driven baselines without physics constraints in both accuracy and stability. 
%Among the evaluated models, the Physics-Guided AFNONet achieves the highest fidelity, effectively capturing long-range spatial dependencies while adhering to physical laws. 
This framework enables computationally efficient,
% near real-time, 
physically plausible fire forecasting to support adaptive prescribed burn management.

\end{abstract}

% \begin{IEEEkeywords}
% Wildfire Modeling, Physics-Guided Machine Learning, ConvLSTM, Transformers, ViViT, Fuel Density Prediction, Prescribed Fire Management.
% \end{IEEEkeywords}
%
%
\section{Introduction}
Wildfires in the United States consume millions of acres each year, posing significant threats to ecosystems, human life, and property\cite{huot2021deeplearningmodelspredicting}. To mitigate these risks, fire managers employ prescribed burns — intentional, controlled fires in designated forest areas designed to reduce fuel loads that can cause dangerous wildfires, enhance forest health, and help manage pests\cite{chatterjee2023prescribedmodelingusingknowledgeguided}. However, these burns are only initiated under specific conditions to ensure safety and prevent unintended consequences~\cite{usfs2022gallinas}.
% , as seen in the 2022 Santa Fe National Forest incident
Fire managers rely on process-based models such as FARSITE~\cite{Finney_1998} and QUIC-Fire~\cite{linn2020quic} to predict the behavior and trajectory of prescribed fires under varying weather conditions (primarily wind speed and wind direction), fuel density in the area, and pre-defined ignition conditions. These process-based models simulate physical fire dynamics to assess burn risks, but their computational demands often prevent real-time decision-making~\cite{ALLAIRE2021184, Linn2002FIRETEC}.
%, especially when weather conditions change rapidly.

While purely data-driven deep learning architectures, such as CNNs~\cite{radke2019firecast, cope2021using} and U-Nets~\cite{ZHANG2021112467}, offer computational speedup, they are ill-suited for complex multi-point ignitions and frequently predict non-physical behaviors like spontaneous fuel regeneration. To counteract this, existing physics-guided ML approaches attempt to constrain outputs using partial differential equations (PDEs)~\cite{bottero2020physicsinformedmachinelearningsimulator, ALLAIRE2021184}. However, these rigid mathematical formulations are often highly unstable, fail to converge, or suffer from well-documented gradient flow pathologies~\cite{wang2021understanding, raissi2019physics}. Our proposed surrogate framework addresses these gaps by embedding domain-specific physical constraints as "soft" penalties, providing the guidance of physics without the numerical fragility of strict equation solving.

In this paper, we present a framework to incorporate physics components into deep learning models that can work across  different architectures.  Specifically, we present a comparative analysis with and without physics-guided loss for three deep learning models, namely Convolutional LSTM (ConvLSTM)~\cite{shi2015convolutional}, Adaptive Fourier Neural Operator (AFNONet)~\cite{guibas2022adaptivefourierneuraloperators}, and Video Vision Transformer (ViViT)~\cite{arnab2021vivitvideovisiontransformer}, for the task of spatiotemporal fuel density change.

% In this paper, we present a comparative analysis of different PGML approaches, namely Convolutional LSTM (ConvLSTM)~\cite{shi2015convolutional}, Adaptive Fourier Neural Operator (AFNONet)~\cite{guibas2022adaptivefourierneuraloperators}, and Video Vision Transformer (ViViT)~\cite{arnab2021vivitvideovisiontransformer}.

%% explain in one para what our paper does ? and how it is different from previous work ? basically what are the novel contributions ?

\section{Dataset and Problem Formulation}

% \subsection{Dataset}

We utilize an ensemble of QUIC-Fire simulations of fire spread, measuring fuel density (mass of fire fuel per unit volume) over time in an area. Each simulation runs for 50 seconds on a 300 $\times$ 300 spatial grid with flat grassland. The dataset comprises all possible combinations of 7 wind speed values (ranging from 1 m/s to 15 m/s), 11 wind direction values (from 230 to 330 degrees), 4 distinct ignition pattern types (aerial, inward, outward, strip north). The wind speed and wind direction remain constant across the entire spatial grid and all timesteps. The data contains fuel densities in the range [$0,0.7kg/m^2$]. 
% However, our models are designed to accommodate dynamically varying weather conditions. 
To ensure effective learning, preprocessing steps include normalization of wind speed and wind direction using min-max scaling, 
% transformation of wind direction into sine and cosine components to better capture angular continuity, 
and incorporation of a source map.  The source map is a 2d array that represents fuel density evolution over time at the lowest wind speed (1 m/s), and the same wind direction and ignition pattern as the input data, allowing the model to learn fuel density propagation patterns. The final dataset consists of four feature channels: source map, wind direction, wind speed, and ignition pattern. 

% \subsection{Problem Formulation}

Our objective is to model the evolution of fuel density as a prescribed fire propagates under different environmental conditions. Specifically, we aim to develop a data-driven emulator of QUIC-Fire, using simulated runs as training data. Each training example, indexed by $i$, is represented by an input tensor $X_i \in R^{T \times H \times W \times C}$ where T denotes the number of timesteps in the sequence, H and W define the spatial grid dimensions, and C represents the number of input channels. The corresponding output tensor for fuel density predictions is given by $Y_i \in R^{T \times H \times W \times 1}$. In our setup, wind speed and wind direction remain static across time.
%—they are only provided at the first timestep but are repeated throughout the sequence to create meaningful embeddings. 
Ignition patterns and source fuel density maps, on the other hand, evolve dynamically over time. By structuring the dataset in this manner, our approach allows the model to learn complex spatiotemporal interactions between fire behavior and environmental conditions, enabling improved predictions of fuel density evolution in prescribed fire scenarios.

\section{Methods}

% We present three deep learning models designed to capture the complex spatiotemporal relationships between atmospheric variables and fuel densities in wildfire spread prediction. These models are specifically chosen for their ability to handle sequential and spatial data, making them well-suited for modeling the dynamic and spatially correlated nature of wildfire behavior. The models include: ConvLSTM, AFNONet, Video Vision Transformer (ViViT). Each of these models is designed to address the unique challenges of wildfire prediction, such as the nonlinear interactions between atmospheric variables (e.g., wind speed, wind direction) and fuel density, as well as the spatial and temporal dependencies inherent in wildfire dynamics. By using these architectures, we aim to improve the accuracy and robustness of wildfire spread predictions, ultimately aiding in better fire management and mitigation strategies.

We evaluated three deep learning architectures
%—ConvLSTM, Adaptive Fourier Neural Operators (AFNONet), and Video Vision Transformers (ViViT)
%—to capture the non-linear coupling between atmospheric variables and fuel density. These architectures were selected 
to represent distinct spatiotemporal paradigms: \textbf{ConvLSTM}
uses a convolutional neural network (CNN)  for spatial structure and a long short-term memory (LSTM) network for temporal dynamics;
\textbf{ViViT} leverages self-attention to model long-range temporal dependencies; and \textbf{AFNONet} utilizes Fourier domain mixing to efficiently capture global dependencies. 
% which is theoretically aligned with solving the fluid-dynamics partial differential equations (PDEs) governing fire spread\cite{raissi2019physics}.

\subsection{ConvLSTM}

Our ConvLSTM model adapts the architecture proposed by Chatterjee et al. \cite{chatterjee2023prescribedmodelingusingknowledgeguided}. While the references work utilizes a four-layer stacked ConvLSTM with Batch Normalization, we simplify this design to a \textbf{two-layer structure} (64 filters, $3\times3$ kernels) to prevent overfitting on limited simulation data.

We introduce two critical architectural enhancements to the standard ConvLSTM to support fuel density modeling:
\begin{enumerate}
    \item \textbf{Residual Connections:} We attach an auxiliary 3D convolutional head to the first LSTM layer to allow injection of physics-based gradients directly into the intermediate layers via residual connections~\cite{he2015resnet}, preventing gradient vanishing during backpropagation through time.
    \item \textbf{Spatiotemporal Smoothing:} 
    We use a 3D kernel of ($3\times3\times3$) which integrates the information across adjacent time steps and pixels, ensuring temporal continuity in the predicted fire front.
    % Unlike the frame-wise 2D projections used in \cite{chatterjee2023prescribedmodelingusingknowledgeguided}, our final output head utilizes a \textbf{3D Convolution} ($3\times3\times3$ kernel) integrating information across adjacent time steps, ensuring temporal continuity in the predicted fire front.
\end{enumerate}

To maintain numerical stability, we use \textbf{LeakyReLU} ($\alpha=0.1$) 
%and initialize the final convolution biases to a small positive value ($0.1$) 
to prevent the "dead neuron" collapse often observed in sparse data like fire maps. 

\subsection{Video Vision Transformers (ViViT)}
We utilize ViViT \cite{arnab2021vivitvideovisiontransformer}, a transformer-based architecture originally designed for video classification, to model the spatiotemporal evolution of fuel density.
ViViT processes its input by tokenizing the video sequence using tubelet embeddings, which serve as the spatiotemporal representation of the fuel grid in our use case. 
% The model outputs two key representations: pooler\_output – unused for prediction, last\_hidden\_state – contains token embeddings, including a classification token. 

To reconstruct the fuel density map, we added additional processing layers  to our ViViT.  Token embeddings are first mapped back to their spatial regions through a token-to-tubelet reconstruction process: A Multi-Layer Perceptron (MLP) projects each token into its respective tubelet dimensions. The reconstructed tubelets are then reshaped into the predicted tensor, followed by a Conv3D layer, which refines the temporal dependencies, ultimately generating the final fuel density prediction for the next timestep.
% The final output has dimensions [B, $1$, H, W], where: B is the batch size, H, W are respectively the height and width of the fuel density map, $1$ represents the single predicted fuel density channel.
This structured approach allows ViViT to effectively model the spatiotemporal dynamics of fire spread, leveraging transformer-based self-attention for accurate fuel density forecasting.

\subsection{Adaptive Fourier Neural Operator Network (AFNONet)}
% Transformers \cite{vaswani2023attentionneed} have supplanted recurrent architectures like LSTMs in many sequence modeling tasks due to their ability to capture long-range dependencies via self-attention and their high parallelizability. In the vision domain, Vision Transformers (ViT) \cite{dosovitskiy2021imageworth16x16words} treat images as sequences of patches, offering global context awareness that traditional convolutional networks lack. However, the quadratic complexity of self-attention $O(N^2)$ becomes a bottleneck for high-resolution spatiotemporal simulations.
% To address this, we employ the Adaptive Fourier Neural Operator (AFNO) \cite{guibas2022adaptivefourierneuraloperators}, which replaces the self-attention mechanism with Fourier-domain token mixing. This reduces complexity to $O(N \log N)$ while effectively modeling the fine-scale differential physics characteristic of fluid dynamics and fire spread.
% We employ the Adaptive Fourier Neural Operator (AFNO)~\cite{guibas2022adaptivefourierneuraloperators} to address the $O(N^2)$ computational bottleneck of standard Vision Transformers \cite{dosovitskiy2021imageworth16x16words} in high-resolution modeling. 
The Adaptive Fourier Neural Operator Network (AFNONet)~\cite{guibas2022adaptivefourierneuraloperators}, also a transformer-based deep learning model, replaces the self-attention mechanism, which has complexity of $O(N^2)$, with Fourier-domain token mixing, reducing complexity to $O(N \log N)$ while effectively capturing the fine-scale differential physics of fluid dynamics~\cite{pathak2022fourcastnet}.
% \subsubsection{AFNO Sequence-to-Sequence Architecture}

We adapt the standard AFNO architecture for spatiotemporal forecasting by implementing an \texttt{AFNONet\_Seq2Seq} model. While standard AFNO models typically function as autoregressive operators mapping $X_t \to X_{t+1}$, our adaptation leverages 3D tokenization to map a context sequence $X_{in}$ directly to a prediction sequence $Y_{out}$ in a single forward pass.
% \subsubsection{3D Spatiotemporal Embedding}
The input tensor $X \in \mathbb{R}^{B \times C \times T_{in} \times H \times W}$ is processed by a 3D Patch Embedding layer. The volume is partitioned into non-overlapping "tubelets" of size $(time, height, width)$. A 3D convolution projects each tubelet into a latent embedding dimension $D$, effectively compressing local spatiotemporal dynamics into a single token grid. 
% These tokens are summed with learnable position embeddings to retain sequence order.
% \subsubsection{Spatio-temporal Mixing}
The token grid is reshaped to merge temporal and vertical spatial dimensions, allowing the AFNO blocks to process the sequence as a unified field.
% Inside each block, a 2D Fast Fourier Transform (FFT) converts the signal to the frequency domain, where a learned weight matrix performs global mixing of spatial modes. This captures long-range dependencies (e.g., wind-driven advection) efficiently before the Inverse FFT restores the spatial representation.

% \subsubsection{Sequence Decoding and Reconstruction}
% The key innovation for sequence generation lies in the prediction head. Unlike classification heads that reduce dimensionality, our decoding head performs a high-dimensional expansion to reconstruct the spatiotemporal volume.

% For each latent token $z_{i}$, a linear projection layer expands the embedding dimension $D$ back to the pixel space of a single tubelet:
% \begin{equation}
%     \hat{v}_{i} = \text{Linear}(z_{i}) \in \mathbb{R}^{C_{out} \times p_t \times p_h \times p_w}.
% \end{equation}
% Here, $p_t$ represents the temporal patch size, allowing a single token to predict multiple future frames simultaneously. We apply a \texttt{PReLU} activation to capture non-linear fire growth characteristics.

% Finally, we apply a "Pixel-Shuffle" (or Un-patching) operation to rearrange these local tubelet vectors $\hat{v}_{i}$ back into the global grid structure:
% \begin{equation}
%     Y_{out} = \text{Rearrange}(\{ \hat{v} \}) \in \mathbb{R}^{B \times T_{out} \times C_{out} \times H \times W}.
% \end{equation}
% This reconstruction step ensures structural consistency, merging the local predictions into a continuous video sequence of fuel density.

Unlike standard classification heads, our decoding head performs a high-dimensional expansion to reconstruct the full video volume. A linear projection maps each latent token back to the pixel space of its original tubelet: $\hat{v} \in \mathbb{R}^{C_{out} \times p_t \times p_h \times p_w}$.
% We apply a \texttt{PReLU} activation to capture non-linear fire growth characteristics. 
Finally, a rearrangement operation (un-patching) merges these local tubelet predictions back into the global grid structure, yielding the continuous output sequence $Y_{out} \in \mathbb{R}^{B \times T_{out} \times C_{out} \times H \times W}$.

\begin{figure}
    \centering
    \includegraphics[width=\linewidth]{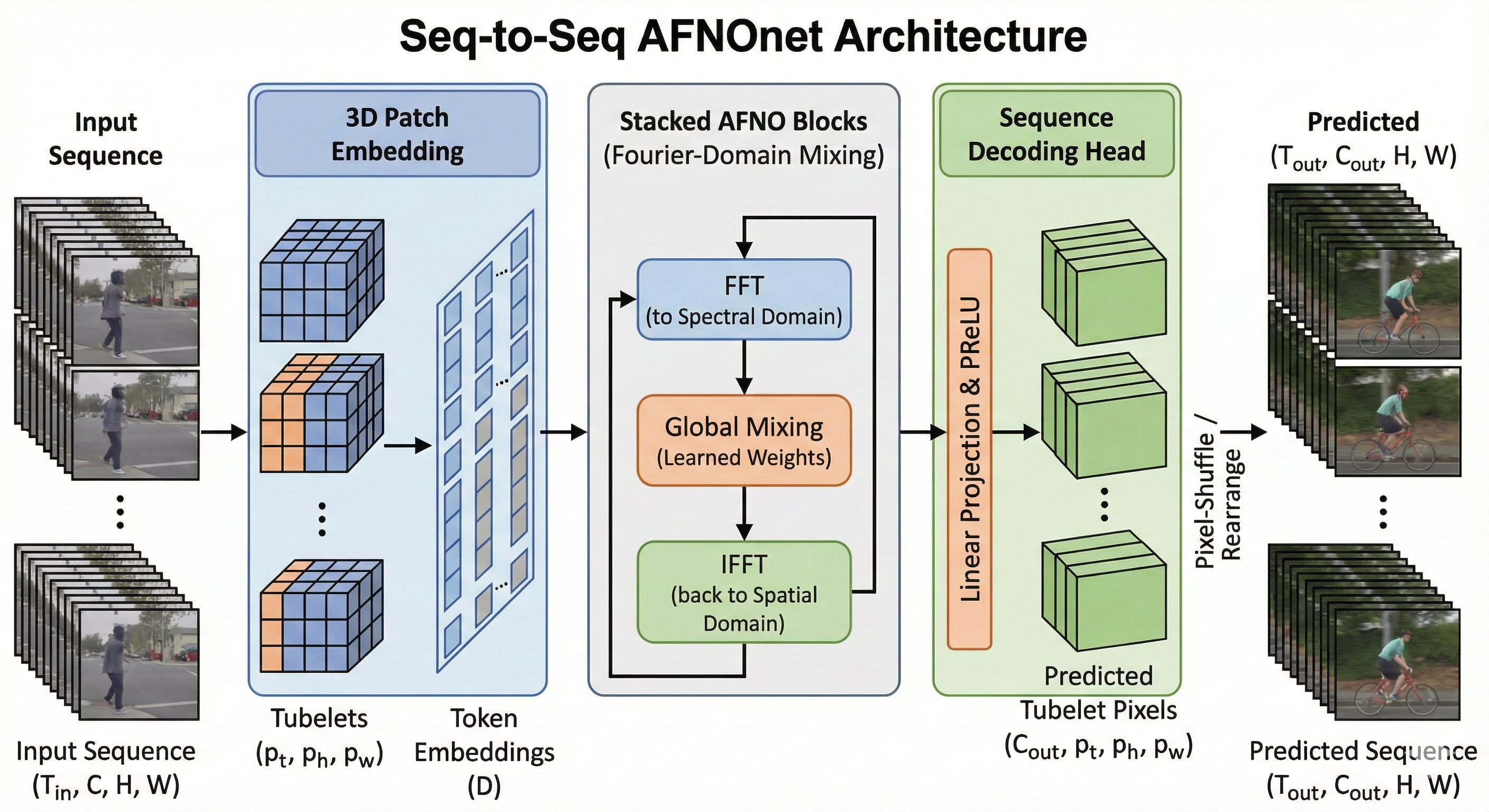}
    \caption{Input frames are tokenized into 3D tubelets and processed via global Fourier-domain mixing to capture long-range dependencies. The decoding head expands latent tokens into dense blocks, applying a pixel-shuffle operation to reconstruct the complete future sequence $Y_{out}$.}
    \label{fig:seq2seq_afnonet}
\end{figure}

\begin{figure*}[t]
    \centering
    \includegraphics[width=\textwidth]{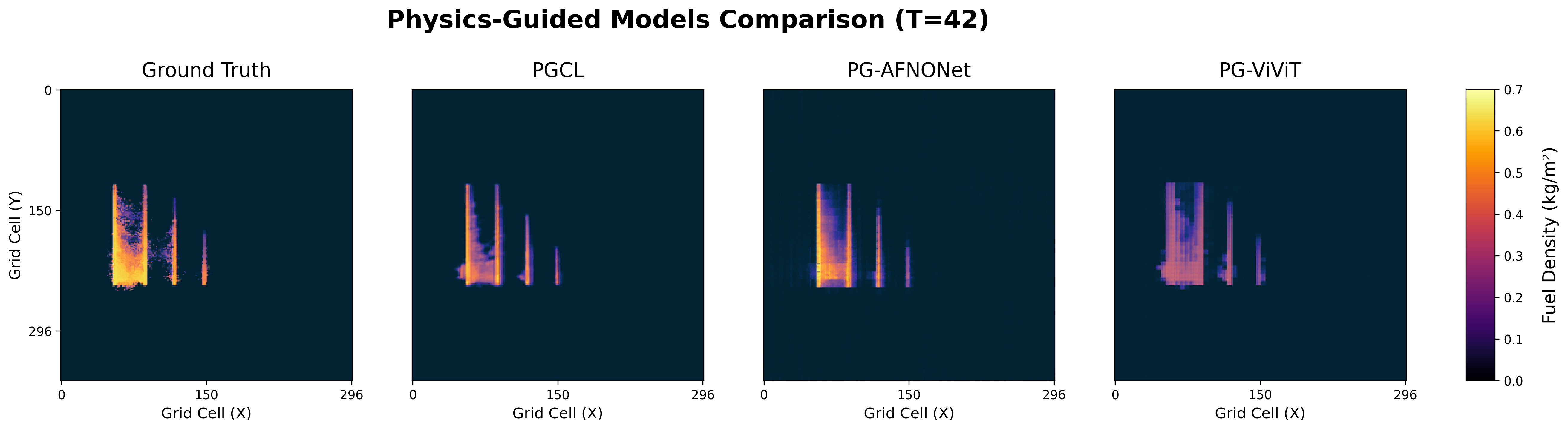}
    \caption{Results of training with physics-guided loss -- Fuel density prediction at timestep 42 for PG-ConvLSTM, PG-AFNONet, and PG-ViViT are displayed, along with ground truth. }
    \label{fig:physics_results}
\end{figure*}

% \begin{figure*}[t]
%     \centering
%     \includegraphics[width=\textwidth]{figures/Fig2_Physics_Fixed.png}
%     \caption{\textbf{Physics-Guided (PG) Results.} Predictions after training with \textit{WiFireLoss}.
%     \textbf{PG-ConvLSTM} utilizes high penalties on state consistency ($\lambda_{burn}=5.0$) to sharpen boundaries.
%     \textbf{PG-AFNONet} balances Fourier mixing with Soft-ROS constraints ($\lambda_{ros}=0.01$), effectively removing spectral noise and achieving the lowest total error.
%     \textbf{PG-ViViT} improves spatial coherence through fuel transport penalties ($\lambda_{trans}=2.0$) but remains limited by patch resolution. All models used a $T_{in}=6 \to T_{out}=3$ horizon.}
%     \label{fig:physics_results}
% \end{figure*}

\begin{table*}[hbt]
    \centering
    \caption{Results of ConvLSTM, AFNONet, and ViViT with and without Physics-Guided Loss Components\\ (Mean and Standard Deviation from 3 Runs)}
    \label{tab:comparison_table}
    \label{tab:results}
    \begin{tabular}{l c c c c c c}
        \toprule
        {\textbf{Model}} & {\textbf{Base MSE}} & {\textbf{Burn}} & {\textbf{Unburn}} & {\textbf{Fuel Trans.}} & {\textbf{ROS}} & {\textbf{Total Loss}}\\ 
        \cmidrule(lr){1-1} \cmidrule(lr){2-7} 
        ConvLSTM         & 5.7e-4 (5.6e-5)            & 2.5e-3 (4.2e-4)  
                         & 3.8e-4 (2.0e-5)            & 1.5e-3 (9.3e-4)   
                         & 8.5e-3 (2.7e-3)            & 5.1e-2 (1.3e-2)\\
        % OLD
        % PG-ConvLSTM      & \textbf{4.5e-4 (6.0e-6)}   & \textbf{1.6e-3 (2.0e-5)}   
        %                  & \textbf{3.4e-4 (8.0e-6)}   & \textbf{2.3e-4 (8.9e-5)}   
        %                  & \textbf{7.4e-3 (5.2e-5)}   & \textbf{3.0e-2 (7.0e-4)}\\
        % \cmidrule(lr){1-1} \cmidrule(lr){2-7}
        PG-ConvLSTM      & \textbf{4.0e-4 (3.1e-5)}                    & \textbf{1.5e-3 (1.4e-4)}   
                         & \textbf{2.9e-4 (2.0e-5)}   & \textbf{2.6e-4 (1.8e-4)}   
                         & \textbf{7.3e-3 (1.2e-4)}   & \textbf{2.7e-2 (2.2e-3)}\\
        \cmidrule(lr){1-1} \cmidrule(lr){2-7}
        % AFNONet          & 5.1e-4 (3.4e-5)            & 1.9e-3 \textbf{(2.0e-5)}   
        %                  & 3.8e-4 \textbf{(1.0e-5)}   & 2.7e-4 (6.0e-6)   
        %                  & 4.3e-3 (6.0e-4)            & 3.2e-2 \textbf{(6.0e-4)}\\
        % PG-AFNONet       & \textbf{4.5e-4 (1.0e-5)}   & \textbf{1.3e-3} (7.0e-5)   
        %                  & \textbf{3.6e-4} (3.0e-5)   & \textbf{7.4e-5 (2.5e-5)}   
        %                  & \textbf{2.7e-3 (3.2e-4)}   & \textbf{2.1e-2} (1.1e-3)\\ 
        \cmidrule(lr){1-1} \cmidrule(lr){2-7}
        AFNONet          & 6.3e-4 (1.8e-5)            & 2.0e-3 (6.0e-5)    
                         & 5.0e-4 (1.0e-5)            & 6.1e-6 (1.0e-6)   
                         & 2.7e-4 (3.2e-4)            & 2.8e-2 (5.0e-4)\\
        PG-AFNONet       & \textbf{5.4e-4 (3.0e-6)}   & \textbf{1.3e-3 (2.0e-5)}   
                         & \textbf{4.7e-4 (3.0e-6)}   & \textbf{1.6e-6 (1.0e-7)}   
                         & \textbf{6.0e-5 (2.7e-6)}   & \textbf{1.7e-2 (3.0e-4)}\\ 
        \cmidrule(lr){1-1} \cmidrule(lr){2-7}
        ViViT            & 7.6e-4 (1.0e-4)            & 3.1e-3 (4.8e-4)  
                         & 5.3e-4 (6.7e-5)            & 7.0e-4 \textbf{(1.4e-4)}    
                         & 5.3e-3 (1.8e-3)            & 5.7e-2 (8.5e-3)\\
        PG-ViViT         & \textbf{7.0e-4 (3.7e-5)}   & \textbf{2.5e-3 (9.0e-5)}  
                         & 5.3e-4 \textbf{(3.9e-5)}   & \textbf{1.6e-4} (1.7e-4)     
                         & \textbf{4.1e-3 (1.3e-3)}   & \textbf{4.3e-2 (4.6e-3) }\\
        \bottomrule
    \end{tabular}
\end{table*}

\section{Integrating Physics Constraints} \label{sec:physics_loss}
To enforce physical consistency, we define a composite loss function composed of the following loss components: 
% To enforce physical consistency, we propose \textit{WiFireLoss}, a composite objective function that aggregates fuel conservation, state consistency, and kinematic elements. The total loss is given by:
% \begin{equation}
%     \mathcal{L}_{\text{total}} = \mathcal{L}_{\text{MSE}} + \lambda_f \mathcal{L}_{\text{fuel}} + \lambda_b \mathcal{L}_{\text{burn}} + \lambda_u \mathcal{L}_{\text{unburn}} + \lambda_r \mathcal{L}_{\text{ROS}},
% \end{equation}

% \subsection{Loss Components}
\vspace{1ex}
\noindent \textbf{1. Base MSE ($\mathcal{L}_{\text{MSE}}$):}
We employ standard pixel-wise Mean Squared Error(MSE) as an objective to minimize the discrepancy between the ground truth and prediction.

\vspace{1ex}
\noindent \textbf{2. Fuel Transport ($\mathcal{L}_{\text{fuel}}$):} Since fuel consumption is irreversible, we penalize non-physical fuel regeneration (i.e., positive temporal gradients) by taking MSE of pixels with increased fuel density over time.
\begin{equation}
    \mathcal{L}_{\text{fuel}} = \frac{1}{N} \sum_t \mathbbm{1}_{(\hat{F}_t > \hat{F}_{t-1})} \odot \left\| \hat{F}_t - F^{GT}_t \right\|^2,
\end{equation}
% \begin{equation}
%     \mathcal{L}_{\text{fuel}} = \frac{1}{N} \sum \left\| \text{ReLU}(\hat{F}_t - \hat{F}_{t-1}) \right\|^2.
% \end{equation}
where $N$ is the number of timesteps, $\hat{F}_t$ is the 2d fuel density prediction map at time $t$, and $F^{GT}_t$ is the 2d fuel density map at $t$ obtained from the ground truth.

\vspace{1ex}
\noindent \textbf{3. State-Weighted Losses ($\mathcal{L}_{\text{burn}}, \mathcal{L}_{\text{unburn}}$):} To weight burned and unburned pixels independently, we separated loss components based on a differentiable mask calculated through the temperature-scaled sigmoid function at time $t$ as:
\begin{equation}
    \label{eq:diffmask}
    P^{GT}_t=\sigma\left({\left({F^*} - F^{GT}_t\right)/\mathcal{T}}\right),
\end{equation}
where $F^*=0.665$ is the heuristically determined fuel density threshold from the ground truth fuel densities, such that $F^{GT}_t > F^*$ are the unburned pixels, and $F^{GT}_t\leq F^*$ are the burned pixels. We scale the sigmoid function by $\mathcal{T}=0.02$, narrowing the intermediate values between $0$ and $1$ to resemble a differentiable mask that can backpropagate during training. Utilizing $P^{GT}$, we calculate the burned and unburned loss components through
\begin{align}
    \label{eq:statelosses}
    \begin{split}
        \mathcal{L_\text{burn}}&=\llangle P^{GT}_t\odot L_{MSE}\rrangle\\ \mathcal{L_\text{unburn}}&=\llangle (\mathbbm{1}-P^{GT}_t)\odot L_{MSE}\rrangle
    \end{split}
\end{align}
% The ground truth fuel density for a burned pixel is very high (i.e., close to 1), and conversely, is very low for an unburned pixel (i.e., close to 0). This can thus be used as a soft mask to differentiate between burned and unburned pixels.  
% ground-truth burn probability
% To focus optimization on critical phase transitions without using hard thresholds, we weight the standard pixel-wise MSE ($\mathcal{L}_{\text{map}} = \|\hat{F} - F_{\text{GT}}\|^2$) using the ground-truth differentiable burn probability $P^{\text{GT}}$ (Eq. \ref{eq:burn_prob}):
% \begin{align}
%     \mathcal{L}_{\text{burn}} &= \text{mean}\left( P^{\text{GT}} \odot \mathcal{L}_{\text{map}} \right), \\
%     \mathcal{L}_{\text{unburn}} &= \text{mean}\left( (1 - P^{\text{GT}}) \odot \mathcal{L}_{\text{map}} \right).
% \end{align}
This formulation ensures that $\mathcal{L}_{\text{burn}}$ dominates in active fire regions, while $\mathcal{L}_{\text{unburn}}$ stabilizes the background.

\vspace{1ex}
\noindent \textbf{4. Rate of Spread ($\mathcal{L}_{\text{ROS}}$):} For a fire propagating with eastward wind, the rate of spread (ROS) is the speed at which the leading edge in the $x-$axis of a fire advances over the landscape. We can calculate the leading edge of an eastward-propagating fire at time $t$ through
\begin{equation}
    \label{eq:argmax}
    \nu(t)=\Delta x\cdot \verb|argmax|_j(F_{ij}(t)),
\end{equation}
where $\Delta x=2m$ is the grid size and $\verb|argmax|_j\left(F_{ij}(t)\right)$ is the index of the pixel with maximum fuel density at time $t$. Using the Eq. \eqref{eq:argmax}, we can calculate the ROS through \begin{equation}
    \label{eq:ros}
    \text{ROS}(t) = \frac{1}{t}\left(\nu(t)-\nu(0)\right).
\end{equation}
However, the \verb|argmax| function in Eq. \eqref{eq:argmax} is nondifferentiable, which prevents backpropagation during training. Therefore, we approximate the leading edge using a series of differentiable equations as follows.

We first project the 2D differentiable mask in Eq. \eqref{eq:diffmask} to a 1D transverse profile through $p_j(t) = \max_i P_t^{GT}$ and calculate a location score, $S_j(t) = j + \lambda (p_j(t) - 1)$, to suppress noise from non-burning columns. Through the use of temperature-scaled softmax of the location scores, $\verb|softmax|(S_j(t)/\mathcal{T})$, we calculate an approximate leading edge for the ground truth fuel map at time $t$, $\nu^{GT}(t)=\sum_j j\cdot p_j(t) \cdot f_j(t)$. We then calculate the approximate ground-truth rate of spread, $\text{ROS}^{GT}(t)$, through the approximate leading edge using Eq. \eqref{eq:ros}. For the predicted approximate rate of spread, $\widehat{\text{ROS}}(t)$, we use the predicted fuel density map, $\hat{F}_t$, instead of the ground truth, $F_t^{GT}$, while calculating the differentiable mask in Eq. \eqref{eq:diffmask}. We constrain the kinematic velocity of the fire front by minimizing the error between the predicted differentiable rate-of-spread (ROS) and the ground truth. ROS is calculated through
\begin{equation}
    \mathcal{L}_{\text{ROS}} = \left\| \widehat{\text{ROS}}(t) - \text{ROS}^{GT}(t) \right\|^2.
\end{equation}

\vspace{1ex}
\noindent \textbf{5. Total Loss ($\mathcal{L}_\text{total}$):}
To enforce physical consistency, we propose \textit{WiFireLoss}, a composite objective function that aggregates fuel conservation, state consistency, and kinematic elements. The total loss is given by:
\begin{align}
    \begin{split}
        \mathcal{L}_{\text{total}} &= \lambda_{\text{MSE}} \mathcal{L}_{\text{MSE}} + \lambda_{\text{fuel}} \mathcal{L}_{\text{fuel}} \\ &+ \lambda_\text{burn} \mathcal{L}_{\text{burn}} + \lambda_\text{unburn} \mathcal{L}_{\text{unburn}} \\ &+ \lambda_\text{ROS} \mathcal{L}_{\text{ROS}},
    \end{split}
\end{align}
where $\lambda_{MSE}$, $\lambda_{fuel}$, $\lambda_{burn}$, $\lambda_{unburn}$, and $\lambda_{ROS}$ are the weights associated with the corresponding loss components.

\section{Experimental Setup and Results}

\subsection{Experimental Setup}
% <<<< Describe model input and output >>>>

All models were implemented in PyTorch using \texttt{bfloat16} mixed precision on a single NVIDIA RTX 3090 Ti. Training utilized the AdamW optimizer with a cosine annealing schedule (Peak $LR=10^{-3} \to 10^{-5}$). Models are trained using a context length of 6 frames ($T_{in}=6$) to predict a 3-frame future window ($T_{out}=3$). For example, given an input sequence of timesteps $t_1, ..., t_6$, the model predicts $t_7, t_8, t_9$. Input and output images are $296\times296$ pixels. The temporal stride is 1, so the next input sequence is $t_2,...,t_7$.

\begin{itemize}
    \item \textbf{ConvLSTM:} Trained over 15 epochs with a batch size of 4 and an early stopping patience of 5 epochs. The hidden dimension size is 64.
    \item \textbf{AFNONet:} Configured with a 3D patch size of $8\times8\times3\ (H\times W\times T)$, an embedding dimension of 256, and 4 layers with 4 spatial frequency blocks. Trained for 50 epochs with a batch size of 8 and patience of 10 epochs. 
    \item \textbf{ViViT:} Utilizes 3D tubelet tokens of shape $(t=3,h=4,w=4)$. The backbone consists of 4 layers, 4 attention heads, an embedding dimension of 256, and an MLP hidden dimension of 512. 
\end{itemize}

We use the following loss component weights ($\lambda$): $\lambda_{MSE}=1$, $\lambda_{fuel}=0.01$, $\lambda_{ROS}=0.01$, $\lambda_{burn}=0.1$, $\lambda_{unburn}=0.1$. 
% OLD: For ConvLSTM, we use: $\lambda_{MSE}=1, \lambda_{fuel}=2, \lambda_{ROS}=1, \lambda_{burn}=5, \lambda_{unburn}=5$. 
These $\lambda$ values were chosen to capture the relative contribution of each component (e.g., burned/unburned are less important than MSE, but more important than fuel transport and ROS). After evaluating several sets of values, these provided the best balance between convergence and training stability.

% \begin{table}[]
%     \centering
%     \caption{Caption}   
%     \begin{tabular}{r c c c c c}
%          \toprule
%          \textbf{Model} & $\bm{\lambda_{MSE}}$  & $\bm{\lambda_{fuel}}$ & $\bm{\lambda_{burn}}$ & $\bm{\lambda_{unburn}}$ & $\bm{\lambda_{ROS}}$ \\
%          \cmidrule(lr){1-1} \cmidrule(lr){2-6} 
%          ConvLSTM  &  1.0 & 2.0 & 5.0 & 5.0 & 1.0 \\
%          AFNONet   &  1.0 & 0.01 & 0.1 & 0.1 & 0.01 \\
%          ViViT   &  1.0 & 0.01 & 0.1 & 0.1 & 0.01 \\
%     \end{tabular}
%     \label{tab:lambdas}
% \end{table}

\subsection{Experimental Results}
Table \ref{tab:comparison_table} presents the performance evaluation across different metrics, including base MSE, state-based errors (burned/unburned), fuel transport violation, and ROS error. The results show that integrating the physics-guided \textit{WiFireLoss} improves fuel density prediction performance across the tested architectures. This demonstrates the generality of the approach for physics-guided machine learning and its applicability to different deep learning models. Importantly, the addition of physics loss terms reduces error means as well as standard deviations, indicating improved and more stable prediction performance.

Fig~\ref{fig:physics_results} illustrates the results of training with \textit{WiFireLoss}.  Fuel density prediction at timestep~42 for all three models are displayed, along with the ground truth. These images show that all models are able to predict fuel density with high fidelity.

\section{Conclusion}

This work develops physics-guided spatiotemporal ML approaches to predict and emulate fuel density evolution over time under different wind conditions and varying ignition patterns. The objective of this project is to enable fire managers to adapt their plans to achieve a burn’s intended objective with changing environmental conditions. Our approach uses a loss function with physical constraints to reduce physical inconsistencies in predictions, including fuel transport discrepancies, over-burning in unburned areas, and over-estimating fuel in burned areas. Our framework can be extended to other spatiotemporal models for learning physically consistent fuel densities. We demonstrate that integrating physics-based constraints into ML models improves fuel density prediction for prescribed fire management. 
% We propose using AFNONet and ViViT model for fuel density prediction. Our findings highlight the effectiveness of ViViT in capturing spatiotemporal dynamics while preserving computational efficiency. 

Future work includes evaluating the proposed method on extended-length fire sequences and on diverse fuel types and topography.  
We also plan to investigate incorporating other features and additional physical constraints. 
%extending PGML frameworks to real-time forecasting and incorporating additional features and physical processes. 
% The proposed method can be modified to include other physical laws and physical constraints. For instance, the work can be extended to three-dimensional grids that include height or the landscape topography as a dimension.

\section*{Acknowledgment}
The authors would like to thank the Societal Computing and Innovation Lab (SCIL) at the San Diego Supercomputer Center, University of California San Diego for their support of this study. This work was funded in part by San Diego Gas and Electric (SDG\&E) and NSF Award 2134904.

%%%%%%%%% REFERENCES
\small
\bibliographystyle{IEEEtranN}
\bibliography{ref}

@String(IJCAI = {IJCAI})

@misc{chatterjee2023prescribedmodelingusingknowledgeguided,
      title={Prescribed Fire Modeling using Knowledge-Guided Machine Learning for Land Management}, 
      author={Somya Sharma Chatterjee and Kelly Lindsay and Neel Chatterjee and Rohan Patil and Ilkay Altintas De Callafon and Michael Steinbach and Daniel Giron and Mai H. Nguyen and Vipin Kumar},
      year={2023},
      eprint={2310.01593},
      archivePrefix={arXiv},
      primaryClass={cs.LG},
      url={https://arxiv.org/abs/2310.01593}, 
}

@misc{guibas2022adaptivefourierneuraloperators,
      title={Adaptive Fourier Neural Operators: Efficient Token Mixers for Transformers}, 
      author={John Guibas and Morteza Mardani and Zongyi Li and Andrew Tao and Anima Anandkumar and Bryan Catanzaro},
      year={2022},
      eprint={2111.13587},
      archivePrefix={arXiv},
      primaryClass={cs.CV},
      url={https://arxiv.org/abs/2111.13587}, 
}

@misc{arnab2021vivitvideovisiontransformer,
      title={ViViT: A Video Vision Transformer}, 
      author={Anurag Arnab and Mostafa Dehghani and Georg Heigold and Chen Sun and Mario Lučić and Cordelia Schmid},
      year={2021},
      eprint={2103.15691},
      archivePrefix={arXiv},
      primaryClass={cs.CV},
      url={https://arxiv.org/abs/2103.15691}, 
}

@book{Finney_1998, title={FARSITE: Fire Area Simulator-model development and evaluation}, url={http://dx.doi.org/10.2737/RMRS-RP-4}, DOI={10.2737/rmrs-rp-4}, institution={U.S. Department of Agriculture, Forest Service, Rocky Mountain Research Station}, author={Finney, Mark A.}, year={1998} }

@article{Linn2002FIRETEC,
  author = {Linn, Rodman R. and Reisner, Jon M. and Colman, Jonah J. and Winterkamp, Judith L.},
  title = {Studying wildfire behavior using FIRETEC},
  journal = {International Journal of Wildland Fire},
  volume = {11},
  number = {3-4},
  pages = {233-246},
  year = {2002},
  doi = {10.1071/WF02007}
}

@article{ALLAIRE2021184,
title = {Emulation of wildland fire spread simulation using deep learning},
journal = {Neural Networks},
volume = {141},
pages = {184-198},
year = {2021},
issn = {0893-6080},
doi = {https://doi.org/10.1016/j.neunet.2021.04.006},
url = {https://www.sciencedirect.com/science/article/pii/S0893608021001337},
author = {Frédéric Allaire and Vivien Mallet and Jean-Baptiste Filippi},
}

@misc{bottero2020physicsinformedmachinelearningsimulator,
      title={Physics-Informed Machine Learning Simulator for Wildfire Propagation}, 
      author={Luca Bottero and Francesco Calisto and Giovanni Graziano and Valerio Pagliarino and Martina Scauda and Sara Tiengo and Simone Azeglio},
      year={2020},
      eprint={2012.06825},
      archivePrefix={arXiv},
      primaryClass={cs.LG},
      url={https://arxiv.org/abs/2012.06825}, 
}

@misc{huot2021deeplearningmodelspredicting,
      title={Deep Learning Models for Predicting Wildfires from Historical Remote-Sensing Data}, 
      author={Fantine Huot and R. Lily Hu and Matthias Ihme and Qing Wang and John Burge and Tianjian Lu and Jason Hickey and Yi-Fan Chen and John Anderson},
      year={2021},
      eprint={2010.07445},
      archivePrefix={arXiv},
      primaryClass={cs.CV},
      url={https://arxiv.org/abs/2010.07445}, 
}

@article{ZHANG2021112467,
title = {Learning U-Net without forgetting for near real-time wildfire monitoring by the fusion of SAR and optical time series},
journal = {Remote Sensing of Environment},
volume = {261},
pages = {112467},
year = {2021},
issn = {0034-4257},
doi = {https://doi.org/10.1016/j.rse.2021.112467},
url = {https://www.sciencedirect.com/science/article/pii/S0034425721001851},
author = {Puzhao Zhang and Yifang Ban and Andrea Nascetti},
}

@techreport{usfs2022gallinas,
  title       = {Gallinas-Las Dispensas Prescribed Fire Declared Wildfire Review},
  author      = {{USDA Forest Service}},
  year        = {2022},
  institution = {U.S. Department of Agriculture, Forest Service},
  url         = {https://www.fs.usda.gov/sites/default/files/gallinas-las-dispensas-prescribed-fire-declared-wildfire-review.pdf},
  note        = {Accessed: 2026-01-17}
}

@article{linn2020quic,
  title={QUIC-Fire: A fast-running simulation tool for prescribed fire planning},
  author={Linn, Rodman R and Goodrick, Scott L and Brambilla, Sara and Brown, Michael J and others},
  journal={Environmental Modelling \& Software},
  volume={125},
  pages={104616},
  year={2020},
  publisher={Elsevier}
}

@article{pathak2022fourcastnet,
  title={FourCastNet: A Global Data-driven High-resolution Weather Model using Adaptive Fourier Neural Operators},
  author={Pathak, Jaideep and Subramanian, Shashank and Harrington, Peter and Raja, Sanjeev and Chattopadhyay, Ashesh and Mardani, Morteza and Kurth, Thorsten and Hall, David and Li, Zongyi and Azizzadenesheli, Kamyar and others},
  journal={arXiv preprint arXiv:2202.11214},
  year={2022}
}

@inproceedings{radke2019firecast,
  title={FireCast: Leveraging Deep Learning to Predict Wildfire Spread},
  author={Radke, David and Hessler, Anna and Ellsworth, Dan},
  booktitle={Proceedings of the 28th International Joint Conference on Artificial Intelligence (IJCAI)},
  year={2019},
  pages={4575--4581}
}

@article{raissi2019physics,
  title={Physics-informed neural networks: A deep learning framework for solving forward and inverse problems involving nonlinear partial differential equations},
  author={Raissi, Maziar and Perdikaris, Paris and Karniadakis, George E},
  journal={Journal of Computational Physics},
  volume={378},
  pages={686--707},
  year={2019},
  publisher={Elsevier}
}

@inproceedings{shi2015convolutional,
  title={Convolutional LSTM network: A machine learning approach for precipitation nowcasting},
  author={Shi, Xingjian and Chen, Zhourong and Wang, Hao and Yeung, Dit-Yan and Wong, Wai-Kin and Woo, Wang-chun},
  booktitle={Advances in Neural Information Processing Systems},
  volume={28},
  year={2015}
}

@article{he2015resnet,
  title={Deep Residual Learning for Image Recognition},
  author={He, Kaiming and Zhang, Xiangyu and Ren, Shaoqing and Sun, Jian},
  journal={arXiv preprint arXiv:1512.03385},
  year={2015}
}

@article{wang2021understanding,
  title={Understanding and mitigating gradient flow pathologies in physics-informed neural networks},
  author={Wang, Sifan and Teng, Yujun and Perdikaris, Paris},
  journal={SIAM Journal on Scientific Computing},
  volume={43},
  number={5},
  pages={A3055--A3081},
  year={2021},
  publisher={SIAM},
  doi={10.1137/20M1318043}
}

@inproceedings{cope2021using,
  title={Using Convolutional Neural Networks to Predict QUIC-Fire Outputs},
  author={Cope, Zachary},
  booktitle={AGU Fall Meeting Abstracts},
  volume={2021},
  pages={NH15A--0448},
  year={2021}
}

\end{document}